\documentclass[10pt,twocolumn,letterpaper]{article}

\usepackage{iccv}
\usepackage{times}
\usepackage{epsfig}
\usepackage{graphicx}
\usepackage{amsmath}
\usepackage{amssymb}

\usepackage{float}
\usepackage{booktabs}
\usepackage{enumitem}
\usepackage{breqn}
\usepackage{mathtools}
\usepackage{multicol}
\usepackage{mathtools}

\usepackage[pagebackref=true,breaklinks=true,letterpaper=true,colorlinks,bookmarks=false]{hyperref}

\iccvfinalcopy 

\newcommand{\OURNAME}{UMFuse\xspace}
\newcommand{\INTMAP}{appearance retrieval map\xspace}
\newcommand{\INTMAPSHORT}{ARMap\xspace}
\newcommand{\xhdr}[1]{\vspace{1em}\noindent{{\bf #1.}}}

\newcommand\blfootnote[1]{%
  \begingroup
  \renewcommand\thefootnote{}\footnote{#1}%
  \addtocounter{footnote}{-1}%
  \endgroup
}

\newcommand\Myperm[2][^n]{\prescript{#1\mkern-2.5mu}{}P_{#2}}
\newcommand\Mycomb[2][^n]{\prescript{#1\mkern-0.5mu}{}C_{#2}}

\def\iccvPaperID{11607} 

\ificcvfinal\fi

\begin{document}

\title{\OURNAME: Unified Multi View Fusion for Human Editing applications}

\author{Rishabh Jain\footnote{rishabhj@adobe.com}
\\
MDSR Adobe\\
\and
Mayur Hemani\\
MDSR Adobe\\
\and
Duygu Ceylan\\
Adobe Research\\
\and
Krishna Kumar Singh\\
Adobe Research\\
\and
Jingwan Lu\\
Adobe Research\\
\and
Mausoom Sarkar\\
MDSR Adobe\\
\and
Balaji Krishnamurthy\\
MDSR Adobe\\
}

\maketitle
\ificcvfinal\thispagestyle{empty}\fi

\blfootnote{{*}rishabhj@adobe.com}

\begin{abstract}
Numerous pose-guided human editing methods have been explored by the vision community due to their extensive practical applications. However, most of these methods still use an image-to-image formulation in which a single image is given as input to produce an edited image as output. This objective becomes ill-defined in cases when the target pose differs significantly from the input pose. Existing methods then resort to in-painting or style transfer to handle occlusions and preserve content. In this paper, we explore the utilization of multiple views to minimize the issue of missing information and generate an accurate representation of the underlying human model. To fuse knowledge from multiple viewpoints, we design a multi-view fusion network that takes the pose key points and texture from multiple source images and generates an explainable per-pixel appearance retrieval map. Thereafter, the encodings from a separate network (trained on a single-view human reposing task) are merged in the latent space. This enables us to generate accurate, precise, and visually coherent images for different editing tasks. We show the application of our network on two newly proposed tasks - Multi-view human reposing and Mix\&Match Human Image generation. Additionally, we study the limitations of single-view editing and scenarios in which multi-view provides a better alternative. \textbf{All Datasplits and results would be made public.}
\end{abstract}
\vspace{-2pt}
\section{Introduction}
\label{sec:intro}

Automating person-image editing can be of great value to business applications for advertisements, commercial merchandising, as well as individual creativity. Pose-based image editing is a class of human image editing workflows where the end result depends on an input human pose. These include human reposing, garment virtual try-on \cite{viton, acgpn, chopra2021zflow} etc. Recent methods for person-image editing \cite{hpt-sota,li2019dense,siarohin2018deformable,dong2018soft,balakrishnan2018synthesizing,ma2017pose, DBLP:journals/corr/abs-1904-03379}  have made great strides in producing accurate edit results from a single image of a person. However, they suffer from textural, shape, and color artifacts because a single image may not have sufficient information to produce an accurate rendition of the person in the desired target pose. In this work, we present a novel pose-guided human image generation method that can incorporate multiple images as well as leverage existing single-view reposing models to improve the editing results significantly.

In most contemporary methods for pose-guided human image reposing \cite{hpt-sota,li2019dense,siarohin2018deformable,dong2018soft,balakrishnan2018synthesizing,ma2017pose}, a deep neural network transforms a \textit{source image} of a person into a \textit{target pose} specified as a sequence of keypoints. The network takes a single source image of a person, the pose of the person in that image (\textit{source pose}), and the \textit{target pose} as input, and produces an image corresponding to the person in the target pose. The source image supplies the color and texture information for guiding the generation of the target pose image. However, the target pose can be significantly different from the source pose, such as when a part of the human body or clothing obscured in the original image might appear in the target pose. In such cases, the network must infer those regions of the output from the context. We posit that using multiple source images of the same person is an effective solution to this problem.
This is also practical for real-world applications as companies usually have multiple photos of a human model in distinct poses. However, this approach requires mapping parts of the target image to parts of the source images based on how their textures match up geometrically, which is a non-trivial task. 
In this paper, we present a framework that can extend existing single-view networks for multi-view reposing task by effectively combining data from multiple source images.
We make the following contributions to the field of pose-guided person image editing:
\begin{itemize}[noitemsep]
\item \OURNAME, a novel plug-and-play framework to fuse multi-scale pose and appearance features from different source images for human image editing applications, using:
\begin{itemize}
    \item an \INTMAP for \textit{interpretable} feature fusion predicted using the input source images and the corresponding pose information.
    \item a visibility-informed pre-training task for initialization of the fusion module.
\end{itemize}
\item The task of multi-view human reposing (MVHR, sec. \ref{sec:results}), and a benchmark datasplit based on the DeepFashion \cite{deepfashion} dataset for evaluating the quality of output. \begin{itemize}
    \item We demonstrate its compatibility with two different single-view reposing networks \cite{gfla, jain2022vgflow}.
    \item We also showcase its versatility by combining different fashion components from completely different persons (Mix-and-Match, sec. \ref{sec:ablation}).
\end{itemize}
\item Detailed quantitative and qualitative analysis, comparisons, and ablation studies indicating the effectiveness of the proposed method.  
\end{itemize}
The next section discusses how the proposed method relates to existing methods for human-image editing.
\section{Related work}

\begin{figure*}
    \centering
    \includegraphics[width=\linewidth]{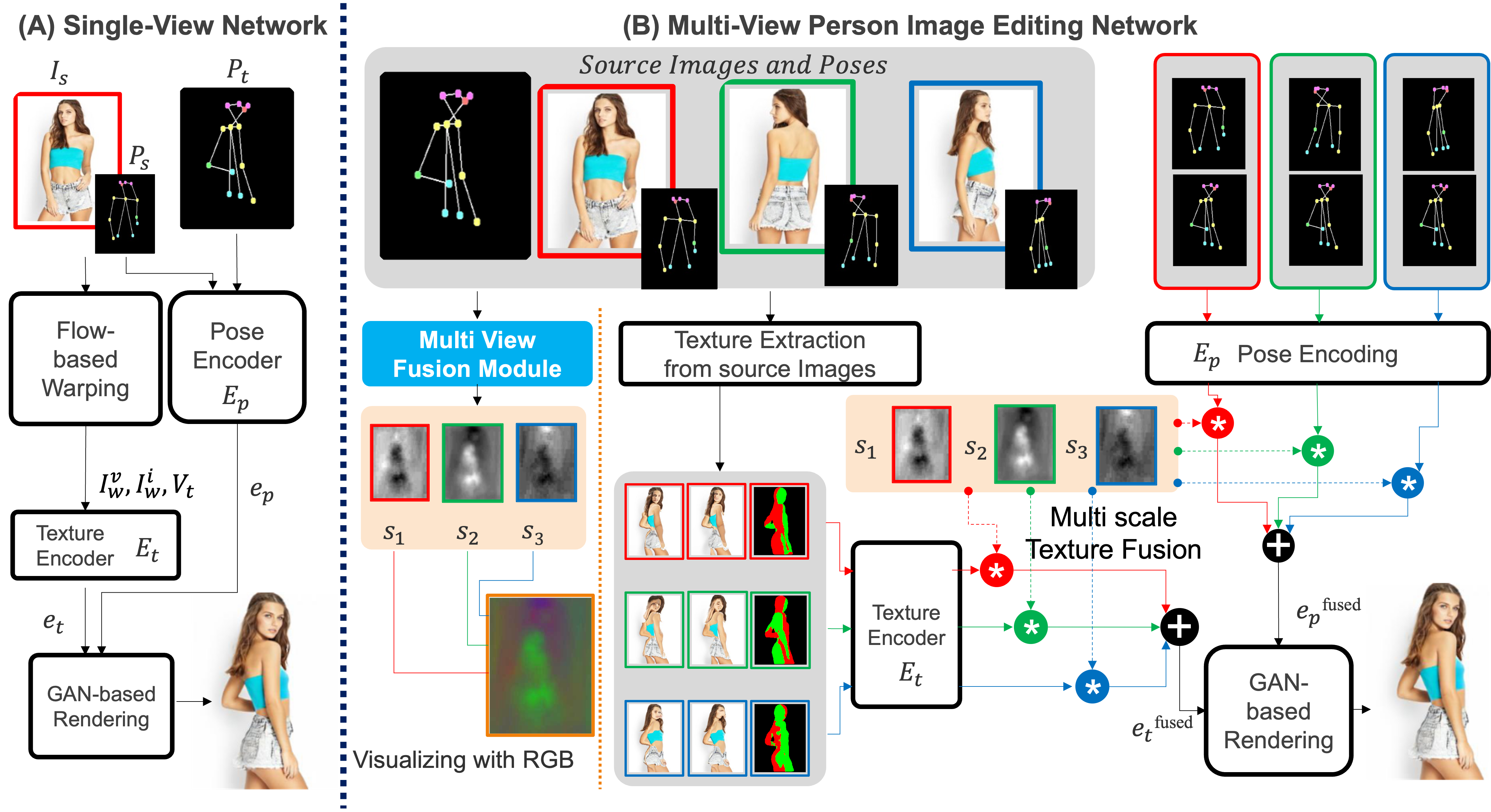}
    \caption{\OURNAME framework: (A) is the single view PHIG network on top of which the \OURNAME framework operates. The source image and its keypoints along with the target keypoints are used to produce warped images and a visibility map ($I_w^v$,$I_w^i$,$V_t$). These are used to obtain the texture encoding at different scales $l$ ($e_{t, l}$), and the source and target poses are used to obtain the pose encoding $e_p$. Together they are used to render the output with a GAN-based renderer. (B) Shows the \OURNAME adaptation for the single view network. The multiple source images and poses are passed individually to the warping and visibility prediction module to obtain multiple warped images and visibility maps, and from those, multiple texture-encoding vectors. Likewise, the source poses paired with the target pose are used to obtain multiple pose encoding vectors. These are merged in an affine combination using the predicted Appearance Retrieval Maps($s_{1-3}$). These maps are obtained using the Multi-view Fusion module and are a key contribution of the \OURNAME framework.}
    \label{fig:umfuse}
\end{figure*}
\paragraph{Pose-guided human image generation} The single view reposing problem has been extensively studied in \cite{siarohin2018deformable, nted,li2019dense,ren2021flow, gfla} . These methods use a single source image as input to produce a reposed person image.
They follow a two-stage strategy -- warping the source image using a flow field to align with the target pose, and  synthesizing the reposed output using a generator. Variations of these methods include leveraging target segmentation mask\cite{spgnet}, visibility maps\cite{li2019dense}, UV-maps\cite{albahar2021pose}, attention-based style distribution\cite{zhou2022cross} etc. for the generation process. While these methods can generate realistic output, they are adversely affected by factors such as heavy occlusions and poor camera angles. We model different body parts in the generated output using multiple source photos where these parts are either clearly visible or have a better style transfer compatibility with the source pose, thereby alleviating the problems faced by single-view methods.

\vspace{-4mm}
\paragraph{NeRFs and Motion Transfer} Recent work in neural radiance field (NeRF) based models \cite{weng2020vid2actor, weng2022humannerf, jiang2022neuman, ramon2021h3d,yenamandra2021i3dmm, gafni2021dynamic, liu2021neural} can reconstruct an entire 3D scene with humans using a relatively larger set of images from one or more videos. Some of them can be adapted to generate novel poses as demonstrated by \cite{jiang2022neuman, weng2020vid2actor}. Few-shot human motion transfer task\cite{huang2021few, wang2019few} also aims to transfer the motion of one person to another via continual frame synthesis. However, apart from requiring a vast number of frames and expensive training, these methods fail to preserve fine-grained clothing texture and produce a coarse body shape output. Also, most of them can reproduce novel poses for only the person they are trained on. Moreover, in contrast to single-view reposing networks, the NeRF-based models have a limited ability to predict the missing content. Our work bridges these two approaches and combines the advantages of using multiple source images with the warping and inpainting capabilities of single-view reposing architectures.

The next section details how \OURNAME produces high-quality, pose-guided human images from multiple sources.
\section{Methodology}
\label{sec:method}
We first describe single-view, pose-guided human image generation (PHIG) as a generic task, and motivate the need for utilizing multiple views using human reposing as an example. Next, we describe the transformation of the single source PHIG network to a multi-view network utilizing multiple source views with \OURNAME.
\begin{figure}[!htb]
    \centering
    \includegraphics[width=\linewidth]{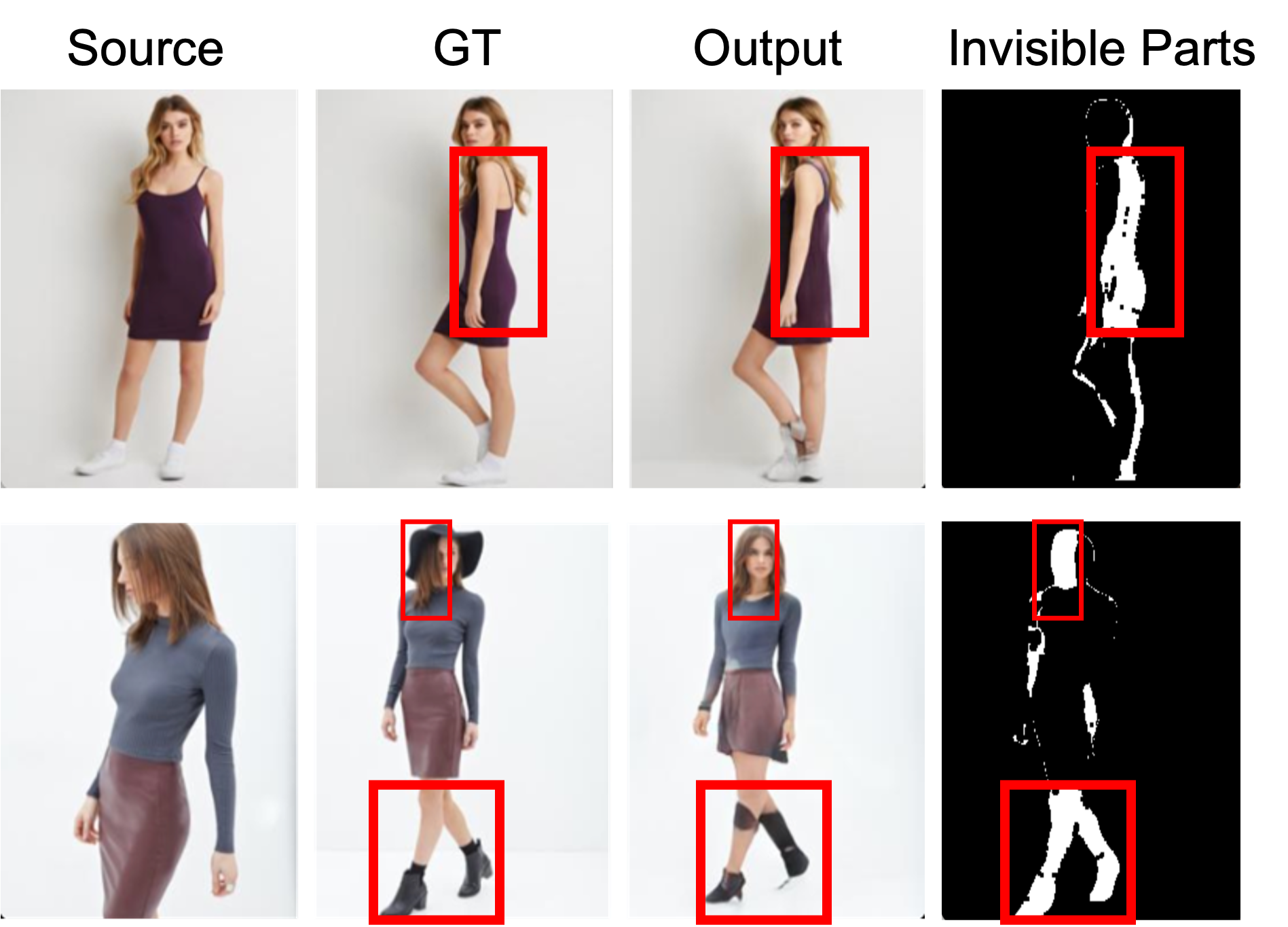}
    \caption{Examples indicating a high degree of overlap between the invisible regions in the visibility maps and the output errors.}
    \label{fig:my_label}
\end{figure}

\paragraph{Single-view Human Image Reposing:} We begin by describing VGFlow\cite{jain2022vgflow}, a recent human reposing network. It broadly consist of three stages (Fig \ref{fig:umfuse}, A) :
\begin{enumerate}[nosep]
    \item Warping the source image $I_s$ to match the target pose $P_t$ and encoding its texture. 
    \item Encoding the source and target pose key points $P_s$ and $P_t$. 
    \item Generation of the final reposed output $I_p$ from both pose and texture encoding. 
\end{enumerate}

The network takes $I_s$, $P_s$, and $P_t$ as input, and predicts two flow fields - $F_v$ and $F_i$. These flow fields are per-pixel 2D displacement fields to warp the source image. Here, $F_v$ represents displacements for regions of the source image that would remain visible when the pose changes to the target. $F_i$ are for the displacements that produce pixels that are invisible in the source pose but may be predicted using context from the source image. These flow fields are used to sample two warped ($I_w$) renditions of the source image -- $I_w^v$ and $I_w^i$ -- corresponding to the visible and invisible regions in the target pose respectively. Additionally, the network also produces a visibility segmentation map $V_t$ of the source image as it would appear in the target pose, indicating the regions that are visible (and invisible) in the source image. Intuitively, the visibility map ($V_t$) is an indicator of whether the network can obtain the correspondence for a pixel from the source image ($I_s$) directly, or it must make an informed prediction.
The encoder stage of the network consists of two modules - the \textit{texture-encoder} ($E_t$), and the \textit{pose-encoder} ($E_p$). The texture encoder captures the color, shape, and style information for the source image $I_s$ at multiple scales($l\ layers$) as it would appear in the target pose by combining information from $I_w^v$, $I_w^i$, and $V_t$. The pose encoder produces a latent representation for the source and target key points ($P_s$, $P_t$) that can be used to condition the generation of a person-image to make it conform to the target pose. 
\begin{align}
\begin{split}
e_{t, l} =& E_t( I_s, P_s, P_t ),  \\
e_p =& E_p( P_s, P_t ) \\ 
\end{split}
\label{eq:ssphig}
\end{align}

Note that \OURNAME is agnostic of the exact architecture used by VGFlow. Other networks like those described in \cite{gfla, albahar2021pose, Cui_2021_ICCV} that have a similar broad architecture are also compatible. We share results for both VGFlow and GFLA \cite{gfla}, in sec. \ref{sec:results}.


The decoder stage ($D$) 
transforms the encoded pose and texture information into an RGB output image $I_p$. One branch of the decoder processes the encoded pose information, while the activations of each layer of the second branch modulate the output of the corresponding layer of the first branch using style modulation\cite{karras2020analyzing} from texture encodings. 

 We take VGFlow\cite{jain2022vgflow} as our primary single-source PHIG network due to its superior performance. Next, we discuss the need for employing multiple inputs for the PHIG task. 
   
\paragraph{Is a single-source input adequate?} Fig \ref{fig:multiViewCompare}  illustrates the limitations of using a single source input image for the human reposing task. We argue that if the target pose is significantly different from the source pose, a PHIG network cannot accurately predict the pixel colors for those pixels that are in the invisible regions. This is expected in extreme cases where one pose is the front, and the other is the back. But even for moderate pose changes, we find that error regions for generated images overlap with the invisible areas in the corresponding ground-truth visibility maps (see Fig~\ref{fig:my_label}). Appearance artifacts can also be caused by non-rigid clothing deformation between poses. Even if the source and target poses are similar in terms of the relative arrangements of the key points, other factors such as the camera position and viewing angle contribute to the errors. Using multiple source images can help improve the output quality if we can learn to map regions in the source views which have the best appearance compatibility with the corresponding target pose regions. Table \ref{tab:multiView-sota} shows that \OURNAME is able to learn this mapping and produces better results by effectively fusing information from up to three source views. We discuss this fusion architecture next.

\paragraph{Combining multiple inputs using an \INTMAP (\INTMAPSHORT):}  \OURNAME employs a feature blending approach for combining multiple source input images. The process is illustrated in Fig \ref{fig:umfuse} (B). Our \textit{Multi-View Fusion(MVF)} module predicts a 2D \INTMAP $s_{1:k}\in [0,1]^{H \times W} $, for each of the $k$ source images. The input to this module is a channel-wise concatenation of the source images ($I_s^{1:k}$), their corresponding key-point representations ($P_s^{1:k}$), and the target key-point representation ($P_t$). The output is a 2D mask with as many channels as the number of input images. 

Hence, the combined soft-selection volume has dimensions $H\times W\times k$. For the present work, we fix the number of input source images to three(k=3). Intuitively, each $(row, col)$ position of the combined soft-selection volume represents the probability of deriving the output pixel at that position from one of the $k$ source images. 

The problem of predicting such a per-pixel \INTMAPSHORT is posed as a joint, conditional soft attention over the input images   
(Eqn \ref{texop}), where the conditioning is on the target pose. We model this as an \INTMAPSHORT prediction task to ensure that the produced weights represent the joint probability of the output pixel being derived from a source image. To this end, we use a network adapted from the Swin Transformer \cite{liu2021swin} with a UPer \cite{xiao2018unified} head. As described in \cite{liu2021swin},  the Swin transformer captures the inter-channel relationships in the input (i.e. between the input images and key-point representations) using self- and cross-attention between shifted windows. Its computational complexity is linear in the size of the output which makes it useful for high-resolution input and output. The UPer head has also been shown in \cite{liu2021swin} to be highly accurate for per-pixel segmentation due to its ability to merge information from multiple scales. The \INTMAPSHORT masks $s_{1:3}$ are obtained by applying \textit{SoftMax} on the output weights obtained from the UPer head at a resolution of 128x128. Formally:

\begin{equation}
s_{1:k} = Softmax(W_{1:k}|I_{s}^{1:k}, P_{s}^{1:k}, P_{t}; \theta)
\label{texop}
\end{equation}

\begin{figure}
\begin{center}
  \includegraphics[width=\linewidth]{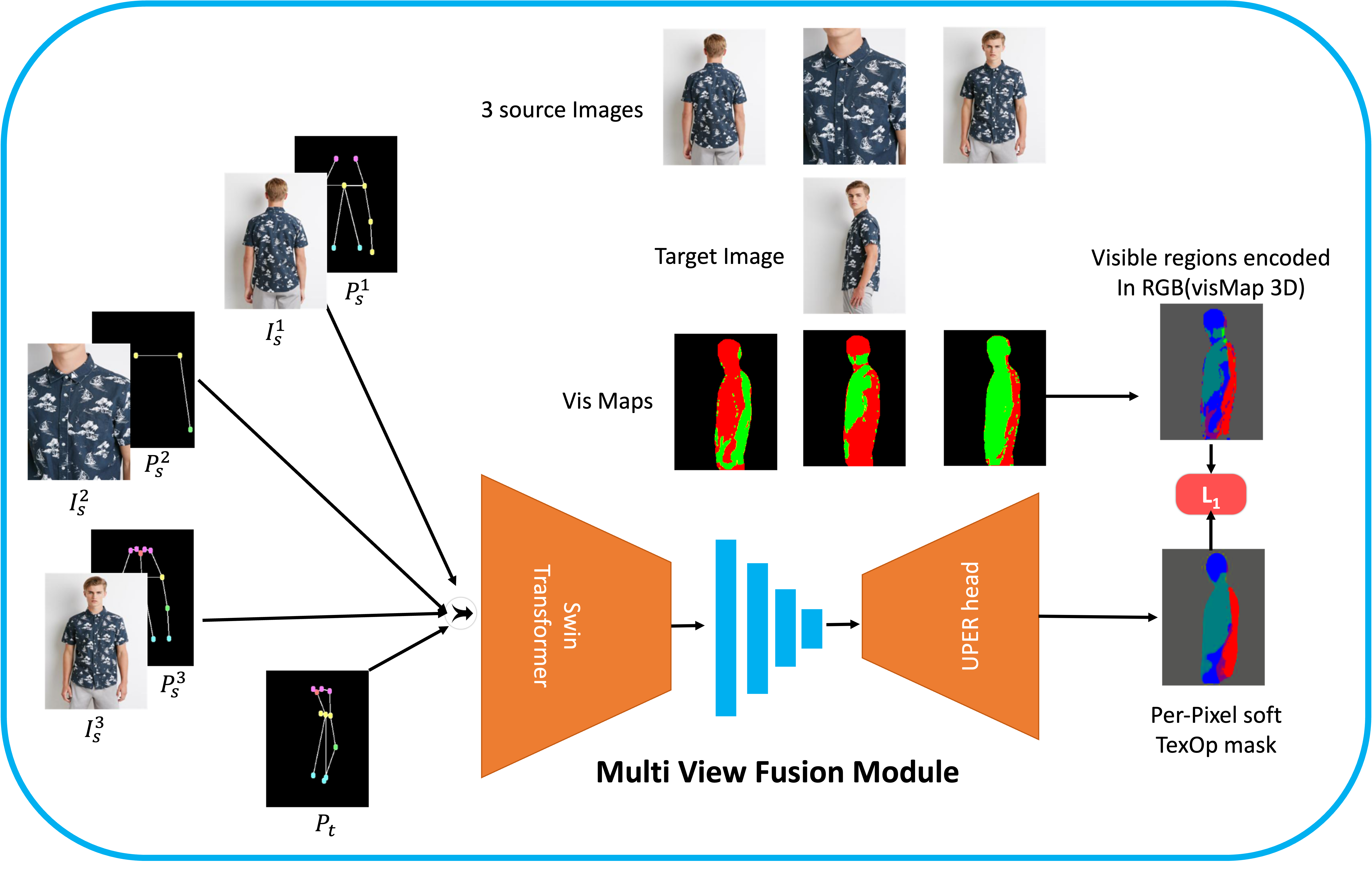}
\end{center}
\caption{Pre-training of our Multi-View Fusion module for learning visibility correspondence with the target pose} 
\label{fig:preTraining}
\end{figure}

Here, $W_{1:k}$ are the computed weights for each of the input images  and $\theta$ are the parameters of our MVF module. The \INTMAPSHORT masks ($s_{1:k}$) are used to fuse the pose- and texture features for the corresponding source images $(I_s^{1:k})$. The fusion is done with the features and not the images themselves to ensure that non-local information present in each image can be combined effectively into the resulting fused features ($e_{t, l}^{fused}$ and $e_p^{fused}$). The computation can be summarized by the following equations:
\vspace{-3pt}
\begin{small}
\begin{align*}
\begin{split}
e_{p}^{fused}; e_{t, l}^{fused} &=  \sum_{\lambda=1}^{k} e_{p}^{\lambda}\odot s_{\lambda} ; \sum_{\lambda=1}^{k}  e_{t, l}^{\lambda} \odot Interp(s_{\lambda}, l)\\[2pt]
\vspace{-1mm}
I_p &=  D( e_p^{fused}, e_{t, l}^{fused} )
\end{split}
\end{align*}
\end{small}

Here, $Interp(s_{\lambda}^{l}, l)$ represents bilinear interpolation of \INTMAPSHORT to the resolution of the $l^{th}$ texture encoding layer and $\odot$ denotes pointwise multiplication of feature encodings. Fig \ref{fig:umfuse} illustrates the entire process.

\vspace{-4mm}
\paragraph{Design \& Pre-Training of fusion module} We essentially wish to model the joint probability of a pixel being derived from a source image using our \INTMAPSHORT. The architectures which match the closest to our desired objective are the ones that were proposed for performing semantic segmentation. The source images can be thought of as different classes and the task of the segmentation network is to predict the likelihood that a certain location should be retrieved from a given input image. However, there are two significant disparities between our intended objective and the output of standard semantic segmentation networks. First, conventional segmentation networks have aligned input and output, allowing them to rely heavily on proximity to establish the class. In our scenario, however, the source humans are all standing in distinct poses. Thus, the network should be able to properly utilize global information from all viewpoints when predicting the probabilities for any pixel. Secondly, unlike semantic segmentation where the task is to predict one class per pixel, our use case requires a prediction of multi-class probability distribution to facilitate blending from different views. Guided by these considerations, our choice of architecture is to use Swin Transformer\cite{liu2021swin} backbone with UPer Head\cite{xiao2018unified} and in Section \ref{sec:ablation} we show comparisons with other alternatives.

To improve its attention to visible regions in each input image (as per the target pose), the MVF module is first trained to predict the visibility map (Fig \ref{fig:preTraining}) using an L1 loss against ground truth generated using a separate 3D pose-estimation(visMap 3D) and projection mechanism based on DensePose \cite{densepose}. We analyze the effect of this pre-training in Section \ref{sec:ablation}.
Finally, the pre-trained network is trained end-to-end for the human reposing task, using the following training objective:
\begin{enumerate}[nosep]
    \item Minimizing the following distances between the output and ground-truth images ($I_p$ and $I_{gt}$):
    \begin{itemize}[nosep]
        \item The pixel-wise mean of absolute difference ($L_1$) for the exact pattern and shape reproduction.
        \item VGG-features based perceptual difference ($L_{vgg}$)
        \item Style difference measured using the mean-squared difference between the Gram-matrices as described in \cite{gatys} ($L_{sty}$).
    \end{itemize}
    The perceptual and style losses help preserve semantic features taken from the input images, such as the identity of the person, and the garment style.
    \item Minimizing an adversarial loss based on LSGAN \cite{LSGAN} ($L_{adv}$) for the output image. This is useful to render realistic output, especially for regions where the decoder must guess the pixel colors. 
\end{enumerate}
The total loss is defined as:
\begin{align*}
    \begin{split}
        L(I_p, I_{gt}) =& \alpha_{rec} \|I_p,I_{gt}\|_1  + \alpha_{per} L_{vgg}(I_p,I_{gt}) \\
&\enskip + \alpha_{sty} L_{sty}(I_p,I_{gt}) + \alpha_{adv} L_{adv}(I_p,I_{gt})
    \end{split}
\end{align*}
where $\alpha_{rec}$, $\alpha_{per}$, $\alpha_{sty}$, and $\alpha_{adv}$ are hyperparameters used to combine the losses.

\begin{figure*}[!t]
\begin{center}
  \includegraphics[width=0.8\linewidth]{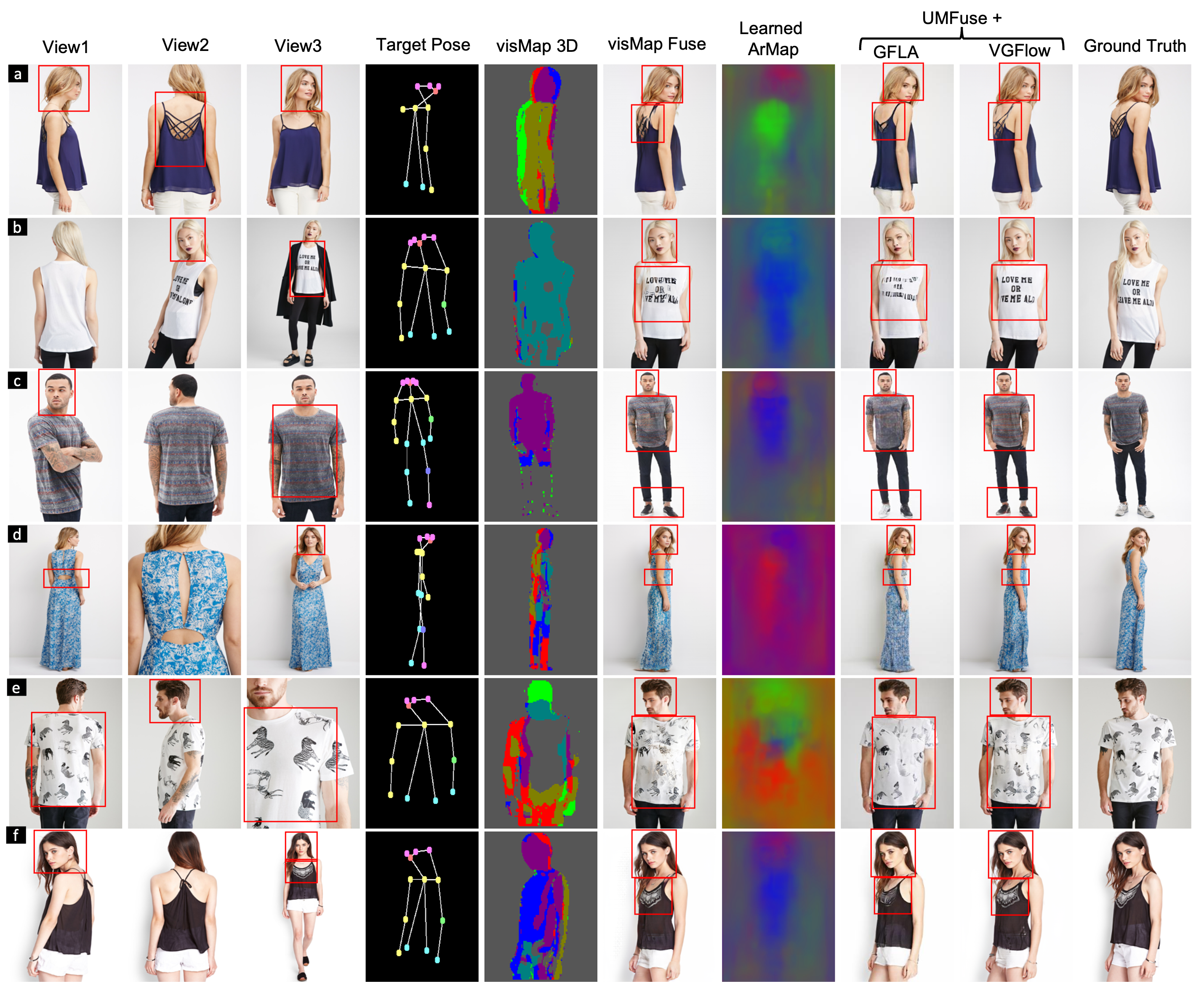}
\end{center}
\caption{Qualitative Results for Multi-View reposing capturing
improvements in (a) handling occlusions, (b) text preservation, (c) inpainting missing information, (d) preserving cloth designs, (e) style transfer from invisible regions, and, (f) pattern reproduction} 
\label{fig:qual1}
\end{figure*}

We observe that the Multi-View Fusion module produces accurate region-specific ArMaps (see Fig \ref{fig:qual1}), attributing the information required to produce accurate and realistic output images. This provides a way to explain the operation of the network as a blending of image regions for reproducing the correct output image. In the following section, we offer a series of experiments and ablations that demonstrate the efficacy of our technique.

\section{Experiments}
\label{sec:exper}

\xhdr{Dataset} We carry out our experiments using the Deepfashion dataset \cite{deepfashion} which has a total of 52,712 high-quality person images with plain backgrounds. Following recent human reposing works \cite{gfla, Cui_2021_ICCV, albahar2021pose, zhou2022cross}, we split the dataset into 48674
training images and 4038 testing images. For all our experiments, we resize the images to 256$\times$256 resolution. We create new training and test quadruples consisting of 3 input source images and 1 output image. We wanted our network to be able to perform reposing task for any number of images $\leq 3$. For training, we utilize all permutations of input source images. We also repeat single source image multiple times in the input to ensure that the network works well if the number of input images are less than three. Note that if a single source image is repeated three times, \OURNAME essentially turns into an identity operator and the network functionally becomes the same as its underlying backbone(GFLA\cite{gfla}, VGFlow\cite{jain2022vgflow} etc). The total number of training samples becomes($	\Sigma\Myperm[n-1]{1} + \Myperm[n-1]{2} + \Myperm[n-1]{3}$), where n is the number of images of a single model in different poses and the sum is over the whole dataset. One of the advantages of performing multi-view training is the polynomial increase in dataset tuples compared to single-view reposing where the samples are limited to $\Sigma\Myperm[n-1]{1}$. The testing split consists of one entry for a distinct 3-view combination($\Sigma\Mycomb[n-1]{3}$) to consistently evaluate the performance of our model for different number of source images(Sec \ref{sec:results}). The order of the testing quadruples was randomly chosen to ensure fairness. Finally, we are left with 720,949 training and 5,247 testing quadruples

\xhdr{Evaluation metrics} Following previous works in supervised person image generation learning, we use the Structural Similarity (SSIM) \cite{seshadrinathan2008unifying}, Peak Signal to Noise Ratio (PSNR) \cite{hore2010image}, Learned Perceptual Image Patch Similarity(LPIPS)\cite{zhang2018unreasonable} and Frechet Inception Distance (FID) \cite{heusel2017gans} metrics to evaluate the results of our multi-view human reposing output. PSNR measures the exact reconstruction accuracy compared to the ground-truth image, while SSIM~\cite{setiadi2021psnr} measures fine-grained similarity based on the local contrast, luminance, and structure. LPIPS computes the patch-wise similarity between two images by comparing the AlexNet activations and FID score measures the realness of the images by computing the 2-Wasserstein distance between the latent space statistics of generated and the ground truth data.

\vspace{-4mm}
\paragraph{Implementation details} 
All experiments are carried out on 8 $\times$ 3090 RTX Nvidia GPUs. For the multi-view fusion module pre-training, we train the model for 3 epochs using a batch size of 32, a learning rate of $10^{-4}$, and the Adam optimizer \cite{kingma2014adam}. For the final training of the Multi-view reposing task, we use a batch size of 24, a learning rate of $3\times10^{-4}$, train for 10 epochs and then finetune only the generator.

\section{Results}
\label{sec:results}

\begin{figure*}[t!]
\begin{center}
  \includegraphics[width=0.87\textwidth]{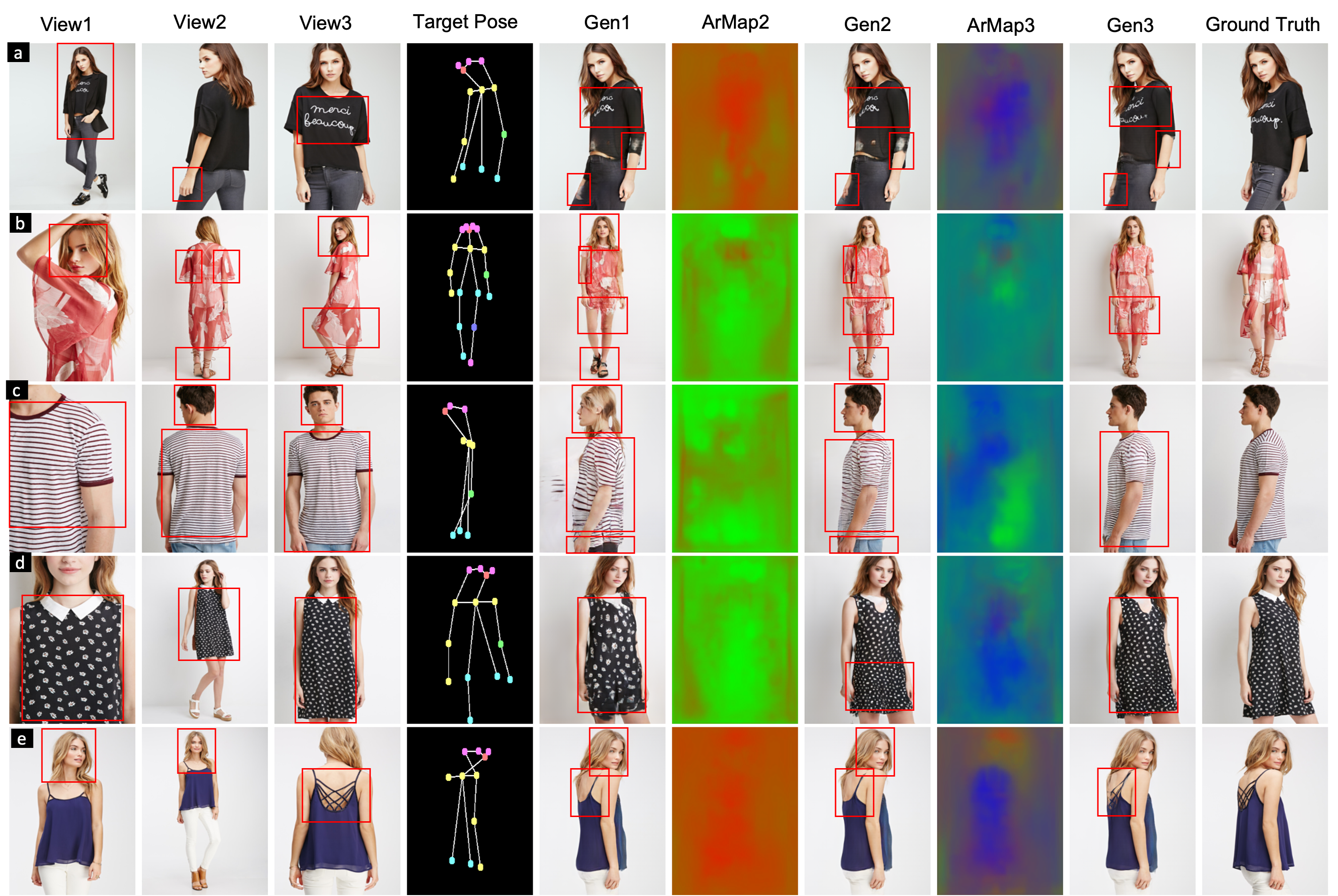}
\end{center}
  \caption{Here, we show incremental improvements gained by the network after the addition of multiple images. Gen1 is generated using only View1 (repeated thrice). Gen2 is generated using View1 and View2 pair with  ArMap2 showing the attribution map using red and green colors respectively. Gen3 is obtained using all the 3 images and ArMap3 encodes the weights in RGB respectively.}
\label{fig:multiViewCompare}
\end{figure*}

Our experiments highlight the advantages of using multiple views (MVHR) over the single view reposing task. The test dataset for the multi-view tests is prepared as discussed in Section \ref{sec:exper}, into quadruples of three source images and one target image (pose). We report the qualitative improvements (Fig \ref{fig:qual1}), as well as the enhancement across a wide set of metrics. Table \ref{tab:multiView-sota} presents the quantitative improvements in output quality obtained by using 2 and 3 unique source views over a single source view (Details in Sec \ref{sec:ablation}). All the image quality metrics(SSIM, PSNR, LPIPS and FID) show a consistent increase with increasing the number of views and are easily attributed to our framework's ability to fill in the missing information or find the best texture correspondence by sampling from distinct views of the same person. We also show that using a single image is fundamentally worse compared to 3 views by testing a source pose that is closest to the target(1-view closest) in terms of the Object-Keypoint Similarity score \cite{OKS} (using a rigid pose-alignment step). To show the generalizability of our method, we freeze our multi-view fusion module and train GFLA\cite{gfla} generator using our methodology. The results show clear improvement over single-view reposing as the network becomes capable of combining texture from different viewpoints. For establishing a baseline, we also perform fusion by visMap3D generated using Densepose\cite{densepose}. This approach confuses the network in areas where a single image would be better(Fig \ref{fig:qual1} Row b) or an invisible region can provide better correspondence (Fig \ref{fig:qual1} Row e). The network also suffers from mistakes due to incorrect 3d pose estimation by Densepose\cite{densepose}. An end-to-end learned approach performs significantly better in establishing the best pixel-level correspondence from source images.

\begin{table}[h!]
\begin{center}
\scalebox{0.9}{\begin{tabular}{cccccc}
\toprule
\textbf{Method}    & \textbf{SSIM} $\uparrow$  & \textbf{PSNR} $\uparrow$  & \textbf{LPIPS} $\downarrow$  & \textbf{FID} $\downarrow$  \\
\hline
\multicolumn{5}{c}{VGFlow + UMFuse for $<3$ Views} \\
\hline

1-view  & 0.716 & 17.16 & 0.204  & 13.52 \\
1-view(closest) & 0.729 & 17.88 & 0.181 & 13.15\\
2-view  & 0.730 & 17.96 & 0.179    & 12.16 \\
\hline
\multicolumn{5}{c}{3-View comparisons} \\
\hline
GFLA+UMFuse & 0.731 & 18.21 & 0.176  & 13.48 \\
VGFlow+visMap & 0.731 & 18.16 & 0.177  & 12.23 \\
VGFlow+UMFuse   & \textbf{0.737} & \textbf{18.32} & \textbf{0.168} & \textbf{12.00}\\
\bottomrule
\end{tabular}}
\end{center}
\caption{Multiple views help in improving SSIM, PSNR and LPIPS metrics significantly over a single view.}
\label{tab:multiView-sota}
\end{table}

Qualitatively, we see that providing multiple source views of a person helps in improving the transfer of different body and clothing characteristics to the output. In Fig \ref{fig:qual1}, row a-c shows the mixing of the model face from one image with the back/side garment view from another. This is further confirmed by visualizing the \INTMAPSHORT. To read the RGB mask - the \textit{red} value represents selection weights for the features of the first source image, the \textit{green} value for the second image, and the \textit{blue} value for the third. For example, in row a, the face is derived from the combination of View1 (red channel) and View3 (blue channel) while the back design of the garment is obtained using the second view (green channel). Row b highlights accurate skin generation and conservation of text geometry while generation. Row c shows the network's capability of in-painting the missing portions which are still not visible in all the source views and Row d 
depicts the accurate reproduction of the texture by fusing appearance appropriately from a desirable scale image(view1). Row e captures the appropriate style transfer to the invisible regions and Row f displays proper pattern reproduction.

\section{Ablation Studies and Analysis}
\label{sec:ablation}

\paragraph{Benefits of using multiple views}  We highlight the incremental improvements done by \OURNAME in  Fig \ref{fig:multiViewCompare} and Tab \ref{tab:multiView-sota}. Gen1 shows the generated output for a single image (View1) repeated 3 times as input. Gen2 uses the pair (View1, View2), repeating View1 for the 3 views. Gen3 is produced by using all 3 views. $\INTMAPSHORT1$ is redundant as the final encoding(after merging using softmax weights) would remain the same regardless of the map. $\INTMAPSHORT2$ is re-normalized to encode the weights assigned to View1 and View2 as Red and Green respectively. We see that in (a), better geometric correspondence for text was obtained after using View3. In (b), the model produced the wrong shoes and dress length in Gen1 which were corrected consecutively in Gen2 and Gen3. In (c), because the face was not visible, Gen1 has the image of a woman and Gen2 was able to correct it but the front region of the cloth still has blurred lines. Gen3 produces the closest match to the ground truth. In (d), we see that the zooming of the source image also has a huge impact on the generated output and the best texture spacing is obtained using a pose which is at a similar zoom level from the Target pose. In (e), we see that even though Gen1 \& Gen2 looks realistic, they still lack the wired pattern at the back of model. 

\begin{figure}[!t]
\begin{center}
  \includegraphics[width=0.95\linewidth]{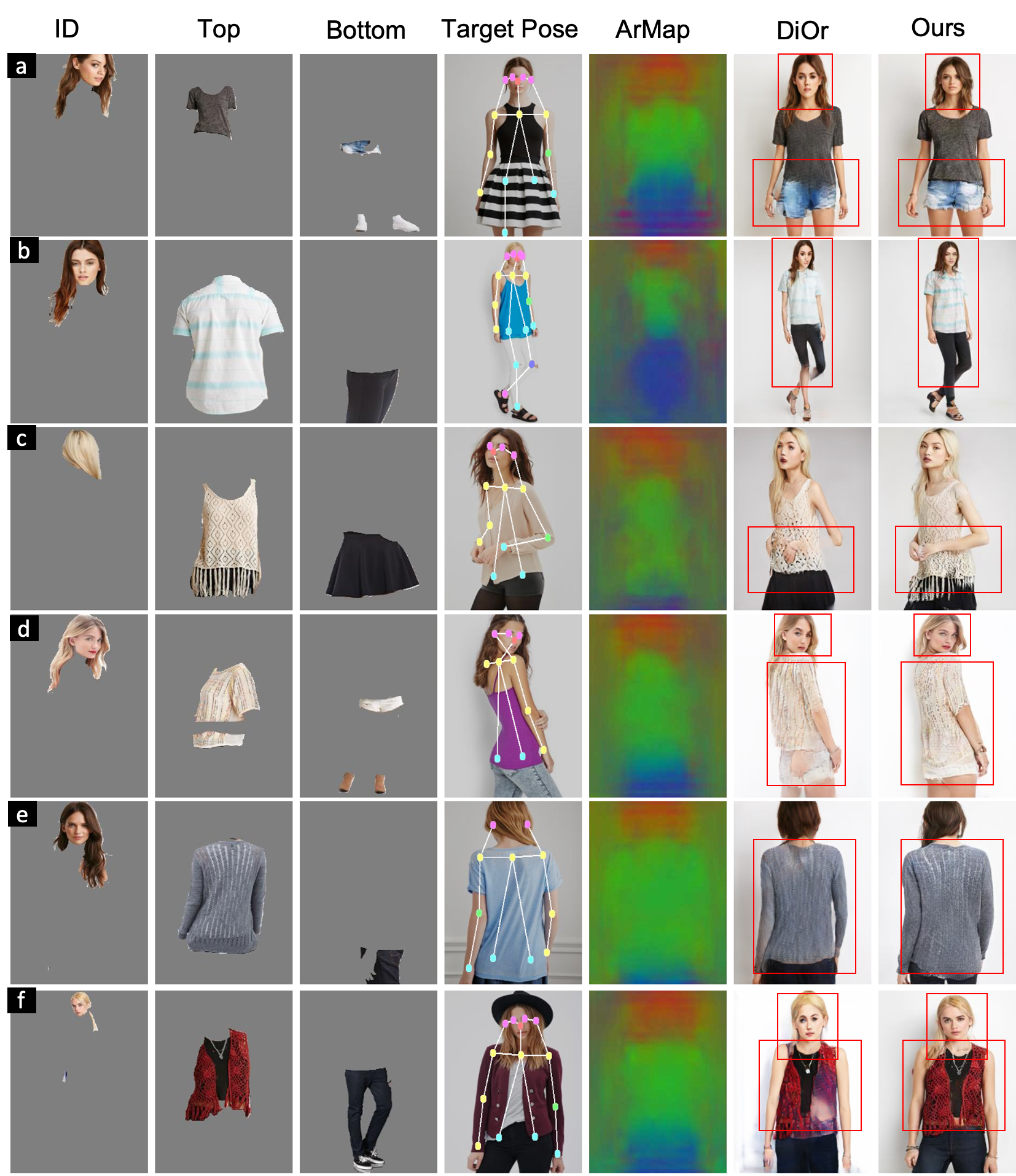}
\end{center}
\caption{Qualitative Results for Mix \& Match Human Image Generation showing realistic generation quality while preserving bodyshape(a), modelling complex pose(b), complicated design and texture of cloth(c), heavy occlusions(d), missing information(e), multiple clothing garments with accessories(f). Additional results can be found in the supplemental material.} 
\label{fig:qual2}
\end{figure}

\vspace{-4mm}
\paragraph{Extension to Mix and match task} Many single view  
reposing models can automatically perform virtual try-on as an additional task\cite{Cui_2021_ICCV, gfla, nted} as there is a natural compatibility between the two. Similarly, multi view reposing  has a natural correspondence with the Mix and match task.  In Mix \& Match Human Image generation(MMHIG), given a tuple (id, top, bottom, pose) from different views, we model all of these jointly to synthesize the final output. We use an off-the-shelf human body parsing network\cite{xie2021segformer} to segment an image into different fashion components($I_{id}, I_{top}, I_{bottom}$) before using the images as input to our network. We then train our network to reconstruct a novel view(guided by $P_t$) which combines all the fashion elements. By learning to disentangle these elements in a supervised fashion, \OURNAME is able to generate accurate rendition of the combined images. More details about the training can be found in supplementary. Note that editing only the pose element will result in single-view human reposing task, Changing top or bottom of the tuple is referred to as virtual try-on and changing id results in identity swapping. In both MVHR and MMHIG tasks, we observe that the Multi-View fusion module produces accurate ArMaps (see Fig \ref{fig:qual1},\ref{fig:qual2}), attributing the information required to produce accurate and realistic output images to the appropriate source views. We compare our output quantitatively and qualitatively to that produced from DiOr\cite{Cui_2021_ICCV} which performs sequential edits on a person image, as it closely matches the Mix-and-Match task. Performing edits sequentially on different parts of the tuple results in inferior output quality as these networks are not able to handle the intersection regions between clothing items and body parts successfully which often results in bleeding and bad texture reproduction. Also, the compounding of errors introduced by these sequential edits result in identity deterioration and inadequate occlusion handling. \OURNAME is a robust, joint learning variant of the error-prone sequential reposing and virtual try-on approach. For quantitative comparison, we prepare 10,000 tuples selected from the test set randomly. We find that \OURNAME significantly outperforms DiOr\cite{Cui_2021_ICCV} with an FID score of \textbf{14.71} vs DiOr’s \textbf{21.73}.

\begin{table}[!h]
\begin{center}
\scalebox{0.8}{
\begin{tabular}{ccccc}
\toprule
\textbf{Fusion Method}    & \textbf{SSIM} $\uparrow$  & \textbf{PSNR} $\uparrow$  & \textbf{LPIPS} $\downarrow$  & \textbf{FID} $\downarrow$  \\
\bottomrule
visMap 3D & 0.731 & 18.16 & 0.177   & 12.23   \\
Swin+UPer(-PT)\cite{liu2021swin} & 0.732 & 18.05 & 0.178 & 12.77  \\
UNet (+PT)\cite{UNet} & 0.735 & 18.30 & 0.172 & 12.37 \\
Swin+UPer(+PT)\cite{liu2021swin} & \textbf{0.737} & \textbf{18.32} & \textbf{0.168} & \textbf{12.00} \\
\bottomrule
\end{tabular}}
\end{center}
\caption{The Swin+UPer(\OURNAME) combination with pre-training(PT)  works best for the reposing task.}
\label{tab:ablation}
\end{table}

\vspace{-6mm}
\paragraph{Multi-View Fusion Module Ablation} In Tab \ref{tab:ablation}, we compare the performance for different methods of latent space fusion. In the first method, we compare against the baseline $visMap 3D$ obtained through Densepose\cite{densepose} to fuse the latent space features. However, direct fusion result in sub optimal performance(Sec \ref{sec:results}). In the second method, we use a Unet\cite{UNet} instead of Swin+Uper combination to predict the selection mask but it wasn't able to produce the desired accuracy due to limited field of vision \& lack of global context in ConvNets. Finally, we ran an experiment of directly training the Swin Transformer without any visibility region Pretraining. As can be seen, there is a significant jump in SSIM(0.732$\to$0.737), PSNR(18.05$\to$18.32), LPIPS(0.178$\to$0.168) and FID(12.77$\to$12.00) metric due to our pretraining. We also show qualitative benefits in supplementary.

\section{Conclusion}
We present \OURNAME - a novel framework for effectively combining information from multiple source images in a pose-guided human image generation pipeline. The approach consists of using a multi-view fusion module that explicitly produces a 2D \INTMAPSHORT for combining features from multiple source images. We demonstrate its effectiveness on two newly proposed tasks: Multi-View human reposing and Mix \& Match Human Image Generation with state-of-the-art results. Our approach can even be utilized as a plug-and-play framework to modify several existing single-view reposing architectures. \OURNAME showcases a promising new and pragmatic approach to produce better results in pose-guided human image generation tasks.

{\small
\bibliographystyle{ieee_fullname}
\bibliography{egbib}
}

\end{document}